\documentclass[journal]{IEEEtran}
\usepackage{amssymb}
\usepackage{amsthm}
\usepackage{graphicx}
\usepackage{multirow}
\usepackage{amsmath,bm}
\usepackage{color}
\usepackage[ruled,vlined]{algorithm2e}
\usepackage{array}

\ifCLASSINFOpdf
\else
\fi
\begin{document}
%
\title{Beyond Triplet Loss: Person Re-identification with Fine-grained Difference-aware Pairwise Loss}
%
%
%

\author{Cheng Yan*, Guansong Pang*, Xiao Bai, Jun Zhou, Lin Gu
\thanks{*Cheng Yan and Guansong Pang contributed equally to this work. Cheng Yan's contribution was made when visiting the University of Adelaide}
\thanks{Xiao Bai is the corresponding author.}
\thanks{Cheng Yan and Xiao Bai are with School of Computer Science and Engineering, Beijing Advanced Innovation Center for Big Data and Brain Computing, Beihang University}
\thanks{Guansong Pang is with School of Computer Science, University of Adelaide}
\thanks{Jun Zhou is with School of Information and Communication Technology, Griffith University}
\thanks{Lin Gu is with National Institute of Informatics, University of Tokyo.}
}

%
%

\markboth{arXiv}%
{Shell \MakeLowercase{\textit{et al.}}: Bare Demo of IEEEtran.cls for IEEE Journals}
%



\maketitle

\begin{abstract}
Person Re-IDentification (ReID) aims at re-identifying persons from different viewpoints across multiple cameras. Capturing the fine-grained appearance differences is often the key to accurate person ReID, because many identities can be differentiated only when looking into these fine-grained differences. However, most state-of-the-art person ReID approaches, typically driven by a triplet loss, fail to effectively learn the fine-grained features as they are focused more on differentiating large appearance differences. To address this issue, we introduce a novel pairwise loss function that enables ReID models to learn the fine-grained features by adaptively enforcing an exponential penalization on the images of small differences and a bounded penalization on the images of large differences. The proposed loss is generic and can be used as a plugin to replace the triplet loss to significantly enhance different types of state-of-the-art approaches. Experimental results on four benchmark datasets show that the proposed loss substantially outperforms a number of popular loss functions by large margins; and it also enables significantly improved data efficiency.  
\end{abstract}


%
\IEEEpeerreviewmaketitle

\section{Introduction}

\begin{figure}[htbp]
\centerline{\includegraphics[width=0.75\linewidth]{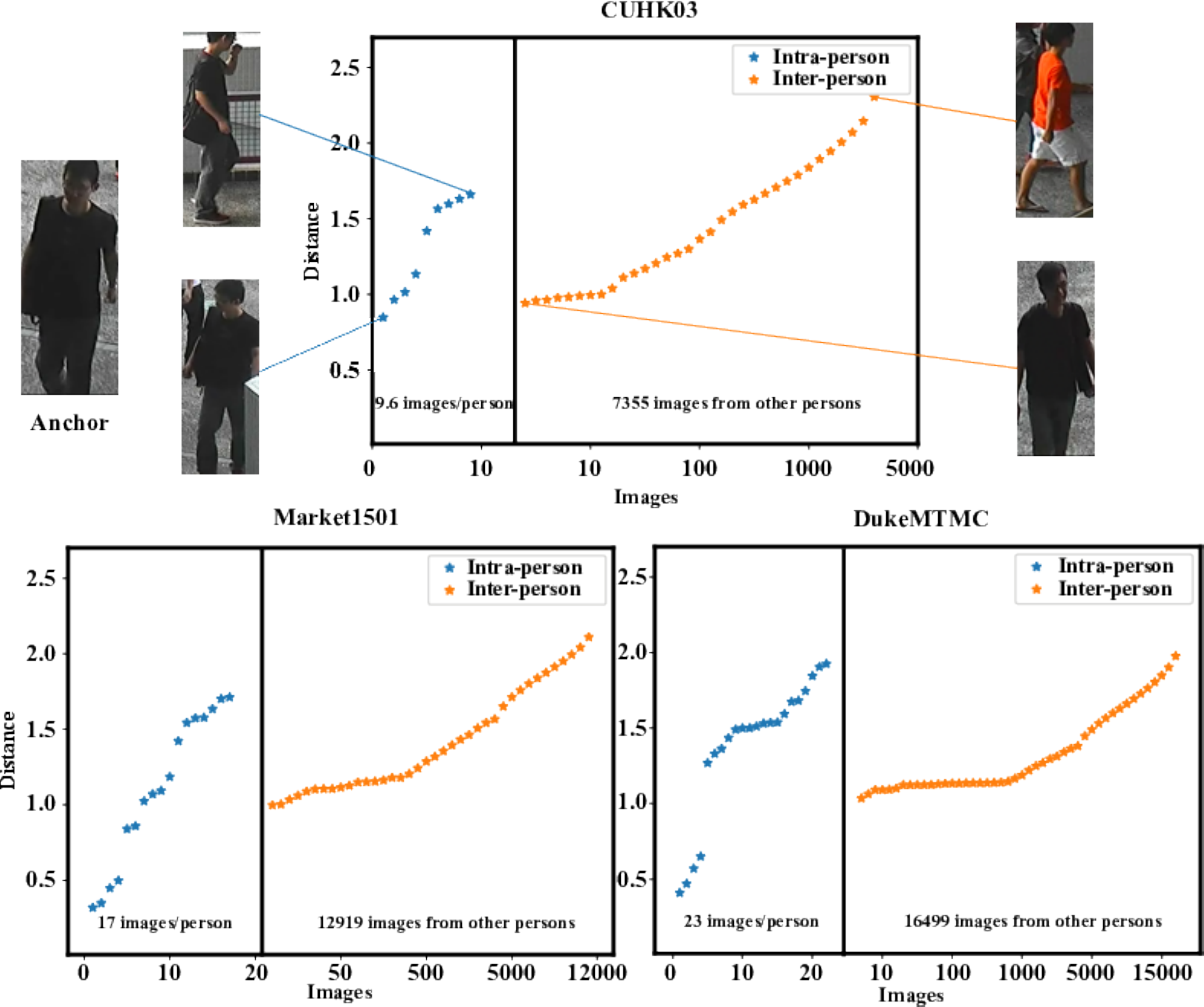}}
\caption{Distribution of average distances between an anchor image and other images from the same person or different persons. Many image pairs have small inter-person distances in popular ReID benchmarks (see Table~\ref{tab:number_of_hard_examples} in Section~\ref{sec:5} for detailed statistics). The distances are calculated using features extracted from ResNet50.}
\label{fig:number of images}
\end{figure}

Person re-identification (ReID), aiming at re-identifying people from viewpoints across multiple cameras, is a critical computer vision task due to its crucial applications in video surveillance, multi-camera tracking and forensic search. Although person ReID has attracted extensive research attentions in recent years, one largely unsolved challenge is how to effectively capture the fine-grained appearance differences of different persons. This problem is crucial to person ReID, because in real-world ReID applications images of different identities can often be differentiated only when looking into these fine-grained differences. This issue manifests itself in popular person ReID benchmarks such as CUHK03~\cite{li2014deepreid}, Market1501~\cite{zheng2015scalable} and DukeMTMC~\cite{zheng2017unlabeled}. To provide a straightforward illustration, we explore and visualize the distribution of average pairwise distances on these datasets. The results are shown in Figure~\ref{fig:number of images}.  It is clear that \textit{inter-person distances}\footnote{
Images of each person are normally treated as samples from an individual class; so class and person/identity are used interchangeable in this study.} (i.e., distance between an image pair of different persons) can be rather small due to fine-grained differences between these images, e.g., the demonstrated CUHK03 anchor image and the negative sample at the right bottom in the first row in Figure~\ref{fig:number of images} have only small differences in the bags and glasses the two persons carry. On the other hand, \textit{intra-person distances} (i.e., distance between an image pair of the same person) can be large due to the fine-grained differences, e.g., the background object in the positive sample at the left bottom in the first row of Figure~\ref{fig:number of images}. Consequently, the identified persons may contain a large number of false positive errors. Similar results can also be observed in the Market1501~\cite{zheng2015scalable} and DukeMTMCC~\cite{zheng2017unlabeled}. Therefore, the ability to capture those fine-grained appearance differences is the key to accurate person ReID.

Inspired by the tremendous success of deep learning, many methods~\cite{su2016deep,liu2017end,wei2018person} have been introduced to learn deep expressive representations for person ReID and achieved state-of-the-art performance. Typically, most of these methods~~\cite{liao2015efficient,xiao2016learning,su2016deep,chen2017beyond,liu2017end,dai2019batch,chang2018multi,luo2019bag,wei2018person,zhong2018camstyle,kalayeh2018human,saquib2018pose,wei2018glad,zeng2020illumination,liu2017provid,luo2020stnreid,xie2020progressive} employ a triplet loss~\cite{liao2015efficient,liu2017end,zhong2018camstyle} or its combination of a classification loss~\cite{dai2019batch,chang2018multi,luo2019bag} as the driving force to extract relevant features. Under this generic framework, several approaches have been developed to learn semantically-rich and/or local features, such as the global feature-based approach~\cite{kalayeh2018human,saquib2018pose}, data augmentation-based approach~\cite{wei2018person,zhong2018camstyle} and striping approach~\cite{sun2018pcb,dai2019batch}.

However, the triplet loss, which enforces that inter-person distances are larger than intra-person distances by a predefined margin, is less effective in learning the fine-grained differences due to two main reasons: (i) as shown in Figure~\ref{fig:frame}, the triplet loss function is dominated by infinitely increasing penalization on large differences between images of the same identity; (ii) it does not enforce sufficient penalization on the images of small differences. For example, triple loss enforces no penalization on the small intra-person differences and imposes a linearly increased penalization on the small inter-person differences. As a result, the triplet loss can only capture the high-level similarities and differences, and thus, it is ineffective in scenes where the fine-grained differences are the key to person ReID.

To address the aforementioned issues, we propose a novel fine-grained difference-aware (FIDI) loss for person ReID. This fine-grained difference-aware property refers to the capability of our loss in adaptively penalizing small inter-person or intra-person appearance differences. Particularly, the FIDI loss enforces an exponentially large penalization on images of those fine-grained differences while at the same time imposing a bounded penalization on their counterparts, i.e., images of large inter-person or intra-person differences. The exponential penalization drives the model to be sensitive to small differences, while the bounded penalization effectively reduces the bias towards large differences. The resulting models can balance expressive features learned from both large and small differences. Additionally, due to the fine-grained difference-aware property, our loss can also leverage the training data more efficiently than the triplet loss.

A number of studies~\cite{chen2017beyond,hermans2017defense,luo2019bag} have dedicated to exploring loss functions other than the triplet loss function for more effective and/or efficient person ReID. Contrastive loss~\cite{hadsell2006dimensionality} is a well-known pairwise loss that learns features for face recognition or re-identification. However, it has similar weaknesses as the triplet loss. Additionally, the single predefined hard margin in these losses also makes it hard to adaptively penalize distance distributions within different person identities.
Quadruplet loss~\cite{chen2017beyond} equips a quadruplet deep network with quadruplet inputs to replace the triplet loss. However, it is limited to specific network structures and is hard to be extended. Batch-hard triplet loss~\cite{hermans2017defense,luo2019bag} is another widely used person ReID loss that optimizes the margin between the most dissimilar intra-person distance and the most similar inter-person distance in each batch. The batch-hard operation is also explored to improve the contrastive loss for the ReID task~\cite{dai2019batch}. The recently proposed circle Loss \cite{sun2020circle} combines the triplet loss with a softmax cross-entropy loss and re-weights each similarity to highlight the less-optimized similarity scores. However, although its batch-wise loss helps regularize the feature learning, it is built upon the triplet loss and thus exhibits similar behaviors in handling the fine-grained feature issues.

In summary, this paper makes the following four main contributions.

\begin{itemize}
    \item We reveal that the widely-used triplet loss function, arguably currently the most popular ReID loss, has inherent difficulties in handling fine-grained appearance differences. This loss is ineffective in challenging ReID cases where different identities can be only distinguished by the fine-grained differences.
    \item We introduce a novel pairwise relationship-based loss function, termed fine-grained difference-aware (FIDI) loss. This FIDI loss enforces exponentially large penalization on small appearance differences while at the same time imposing bounded penalization on large differences. As a result, the FIDI-enabled models can effectively learn expressive features from both large and small appearance differences.
    \item The fine-grained difference-aware property also empowers the FIDI loss to harness the image samples more effectively and is thus substantially more data-efficient than the triplet loss.
    \item We demonstrate that the FIDI loss can be used as a plugin to replace the triplet loss and work effectively in different types of state-of-the-art approaches.
\end{itemize}

Experimental results on four benchmark datasets show that the FIDI loss substantially improves the triplet loss by a large margin, \textit{e.g.}, typically 10\%-20\% improvement in effectiveness. We also show the FIDI loss based models can also largely outperform state-of-the-art vehicle ReID models.

The rest of our paper is organized as follows: In Section~\ref{sec:2}, we review the related works for person ReID. Then we provide corresponding research background and discuss relevant loss functions for person ReID in Section~\ref{sec:3}. Section~\ref{sec:4} introduces the proposed FIDI loss function. Experimental results, visualization and ablation studies are presented in Section~\ref{sec:5}. Finally, the conclusions are given in Section~\ref{sec:6}.

\section{Related Work}
\label{sec:2}
Many studies~\cite{xiao2016learning,cheng2016person} learn feature representations for person ReID by fine-tuning convolutional networks with a classification loss. Different approaches have been introduced to further improve the performance, including data augmentation, striping, and global feature approaches. In this section, we review three types of person ReID approaches.

\subsection{Data Augmentation-based Approach}
Data augmentation is an effective way to improve the feature learning capacity for CNN. There are generative adversarial networks (GANs)~\cite{wei2018person,zhong2018camstyle}, pose estimation~\cite{kalayeh2018human,saquib2018pose}, random erasing~\cite{zhong2017random} in this category.

GAN based approaches use GAN to generate more data for training. Mask or pose guided frameworks obtain the semantic information from pose estimation or segmentation models. These methods use other networks to generate image to increase the number of input images or improve the mask of input for augmentation. However, the benefit comes from the help of other networks with extra semantic information. By contrast, random erasing randomly selects a rectangle region and assigns random values either on image~\cite{zhong2017random} or CNN feature maps~\cite{dai2019batch}. Among these data augmentation-based methods, random erasing is arguably the simplest yet highly effective method without extra computation cost.

\subsection{Striping-based Approach}
Striping based methods aim at enforcing the learner to pay more attention to different parts of the identities by combining striping local features. Part based networks are widely adopted in these methods~\cite{cheng2016person,sun2018pcb,dai2019batch} to separate feature maps into several parts. They build multi-branch neural networks to learn local features in each of the predefined parts of the identities with one-branch network dedicated to one part, and then they concatenate these features to perform ReID during inference.

PCB~\cite{sun2018pcb} is the first part-based deep learning methods for person re-identification. It replaces the original global pooling layer with a spatial conventional pooling layer to separate the last convolution into several pieces of column vectors for independent pooling, in which each part refers to a body part of person. These feature then are concatenated for final feature learning. To further improve the accuracy, both global feature and part-based local feature are learned and used ~\cite{wang2018learning,dai2019batch}. The added features often result in accuracy improvement.

Though these striping methods are often the best performers on different benchmark datasets, they often involve more network parameters and expensive computation than single-branch network-based methods.

\subsection{Global Feature-based Approach}
Global feature-based approach focuses on  a single network structure as the backbone to learn global identity features. These methods work on the sampling process ~\cite{huang2018adversarially,wang2018mancs}, loss design~\cite{hermans2017defense} or learning process~\cite{luo2019bag}. Among them, the loss function is very crucial for feature learning and most relevant to our work.

The combination of classification loss and ranking loss such as triplet loss is one of the most widely-used loss functions for person ReID ~\cite{liao2015efficient,xiao2016learning,su2016deep,chen2017beyond,liu2017end}. The triplet loss often works better than contrastive loss, since the triplet inputs provides a better guarantee of the distance margin than the pairwise contrastive loss. However, the triplet constraint is loose in the sense that it ignores the triplets when the predefined margin is met. The triplet loss is also cumbersome as the triplet sample space is often excessively large for large-scale data. 
A few studies attempt to address these issues for person ReID. One such example is a quadruplet loss with quadruplet network~\cite{chen2017beyond}, but it is limited to specific network structures. Circle loss \cite{sun2020circle} is another closely related work that re-weights each similarity to highlight the less-optimized similarity scores, but the weighting factors are defined in a self-paced manner and need more calculation.
Other methods~\cite{hermans2017defense,luo2019bag} avoid the explicit generation of hard triplet samples. Instead they work with batch-wise hard triplet loss, which optimizes the margin between the most dissimilar intra-person distance and the most similar inter-person distance in each batch. This enhanced triplet loss becomes more sensitive to small appearance differences than the basic triplet loss. However, its inherent penalization mechanism does not change.

\section{Research Background}
\label{sec:3}
This section introduces person re-identification problem and a widely-used state-of-the-art frameworks to illustrate how our proposed loss could be plugged in.

\subsection{Problem Formulation}
In a person ReID system, let $\mathcal{X}=\{\mathbf{x}_{i}, y_{i}\}^{N}_{i=1}$ be a set of $N$ training samples, where $\mathbf{x}_{i}$ is an image sample and $y_{i}$ is its identity/class label. The person ReID algorithm  learns a mapping function $\phi: \mathcal{X} \mapsto \mathcal{F}$ which projects the original data points $\mathcal{X}$ to a new feature space $\mathcal{F}$. This space $\mathcal{F}$ should shrink the intra-person distance  while push the inter-person distance as large as possible. Given a query image $\mathbf{q}$ and $\phi$, the ReID algorithm first computes this distance between $\phi(\mathbf{q})$ and every image $\phi(\mathbf{x})$ from a gallery image set $G$, and then returns the images that have the smallest distance. It should be noted that, for the sake of real-world applications, the gallery image set and the training image set have no overlapping, \textit{i.e.}, the query person does not appear in the training set. Therefore, is is also regarded as a zero-shot problem. This largely distinguishes person ReID from general image retrieval tasks.

\begin{figure*}[htbp]

\centerline{\includegraphics[width=0.95\linewidth]{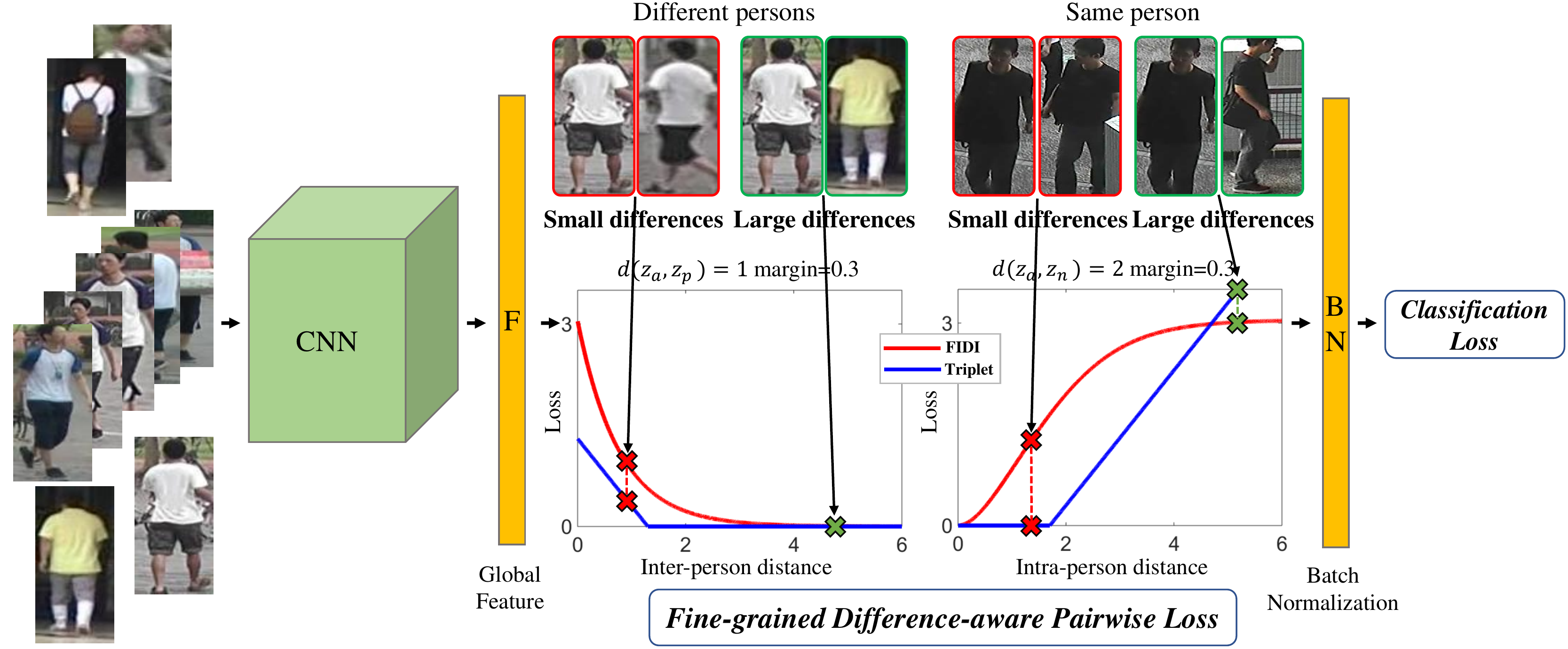}}

\caption{An overview of the fine-grained difference-aware pairwise loss-based framework. It consists of a deep CNN-based network backbone, our proposed FIDI loss and a classification loss. This framework  is exactly the same as the widely-used triplet loss-based framework except that the triplet loss is replaced with our FIDI loss. The network backbone can be various CNN architectures. Unlike the triplet loss that neglects small appearance differences due to the potential dominance of unbounded penalization on images of large intra-person differences, our FIDI loss can effectively capture the fine-grained intra-person/inter-person appearance differences, \textit{e.g.}, image pairs having small appearance difference as in the red boxes above. We achieve this by enforcing exponentially large penalization on images of small differences and bounded penalization on images of large differences.}
\label{fig:frame}
\end{figure*}

\subsection{Triplet Loss-based Approach}
The triplet loss is a widely-used loss function which takes a collection of triplet samples to learn feature representations space where the inter-class distances are greater than intra-class distances by at least a predefined margin $m$. A triplet is composed of three samples $\mathbf{x}_{a}$, $\mathbf{x}_{p}$ and $\mathbf{x}_{n}$, where $\mathbf{x}_{a}$ is an anchor sample. $\mathbf{x}_{p}$ is a positive sample that comes from the same person as $\mathbf{x}_{a}$, while $\mathbf{x}_{n}$ is a negative sample taken from an identity different from that of the anchor. The generic triplet loss (TL) is given as follows:
\begin{equation}
\label{triplet}
L_{tl}=  [d(\mathbf{z}_{a}, \mathbf{z}_{p})-d(\mathbf{z}_{a}, \mathbf{z}_{n}) + m]_{+},
\end{equation}
where $\mathbf{z}=\phi(\mathbf{x})$ denotes the learnt feature representation of $\mathbf{x}$. $d(\cdot,\cdot)$ is the distance of two samples. $m$ is a predefined margin and $[\cdot]_{+}$ represents $\max(\cdot, 0)$. Contrastive loss can be regarded as a special case of triplet loss where $d(\mathbf{z}_{a}, \mathbf{z}_{n}) + m$ is 0 for similar pairs and $d(\mathbf{z}_{a}, \mathbf{z}_{p})$ is 0 for dissimilar pairs. Convolutional networks are often employed to instantiate the $\phi$ function. The triplet loss is the key ingredient here, but Eqn.(\ref{triplet}) requires the  high-quality triplets as input. An advanced triplet loss, termed batch triplet loss (BTL) that is widely-used in person ReID, incorporates hard triplet mining into the loss calculation in each batch~\cite{liao2015efficient,su2016deep,chen2017beyond,liu2017end}. BTL is defined as follows:
\begin{equation}
\label{triplet_hard}
L_{btl}=  [\max_{p=1...B_{p}} d(\mathbf{z}_{a}, \mathbf{z}_{p})- \min_{n=1...B_{n}} d(\mathbf{z}_{a}, \mathbf{z}_{n}) + m]_{+},
\end{equation}
where $\max_{p=1...B_{p}} d(\mathbf{z}_{a}, \mathbf{z}_{p})$ represents the maximum distance between anchor and all $B_{p}$ positive samples in a batch. $\min_{n=1...B_{n}} d(\mathbf{z}_{a}, \mathbf{z}_{n})$ represents the minimum distance between anchor and all $B_{p}$ negative samples in the batch.

To complement the triplet-based local features, a classification loss is used in recent methods~\cite{wang2018learning,dai2019batch,luo2019bag} to work together with the triplet loss for a global constraint in the optimization. This helps   learn class-level global features effectively. The classification loss is defined as:
\begin{equation}
\label{cls}
L_{cla} = \sum_{i=1}^{N} \mathbb{E} (\mathbf{z}_{i}^\intercal \mathbf{W}, \mathbf{y}_{i}),
\end{equation}
where $\mathbb{E}(\cdot)$ is the cross entropy loss. $\mathbf{W}$ is the weight matrix to map $\mathbf{z}_{i}$ to classification labels $\mathbf{x}_{i}$ encoded as one-hot vector $\mathbf{y}_{i}$  in the output layer. This classification loss is added in the output classification layer. A batch normalization layer is normally employed between the triplet loss-enabled feature layer and the output layer to speed up training and stabilize the performance.

This framework works effectively in different benchmark datasets. Recent advances incorporate data augmentation or striping strategies~\cite{wang2018learning,dai2019batch} to achieve new state-of-the-art performance. However, the triplet loss, either $L_{tl}$ or $L_{btl}$, fails to learn expressive features from fine-grained differences. This is because: (i) the triplet loss is not sensitive to small differences, i.e., it enforces no penalization on small intra-person differences or small penalization on small inter-person differences; (ii) the loss grows linear infinitely w.r.t. the increasing intra-person distances and has no upper bound. As a result, the optimization may be dominated by large intra-person differences.


\section{Fine-grained Difference-aware (FIDI) Loss}
\label{sec:4}
This section introduces our fine-grained difference-aware (FIDI) loss to address the bottleneck issue with the triplet loss.

\subsection{The Proposed Framework}

Our proposed framework aims to leverage the capability of the FIDI loss in capturing fine-grained differences to learn well discriminative and generalized features for the person ReID task. Specifically, as shown in Figure~\ref{fig:frame}, our framework is composed of three modules: deep convolutional network-based feature mapping, the FIDI loss and the classification loss. We use exactly the same framework as the triplet loss approach except that the triplet loss is replaced with our FIDI loss. Note that we use this setting to facilitate a straightforward comparison with triplet loss-based approaches in our empirical studies. As discussed in Section~\ref{subssec:striping}, The FIDI loss could also improve other frameworks.

\begin{figure*}[htbp]
\begin{minipage}[b]{0.32\linewidth}
  \centering
  \centerline{\includegraphics[width=0.95\linewidth]{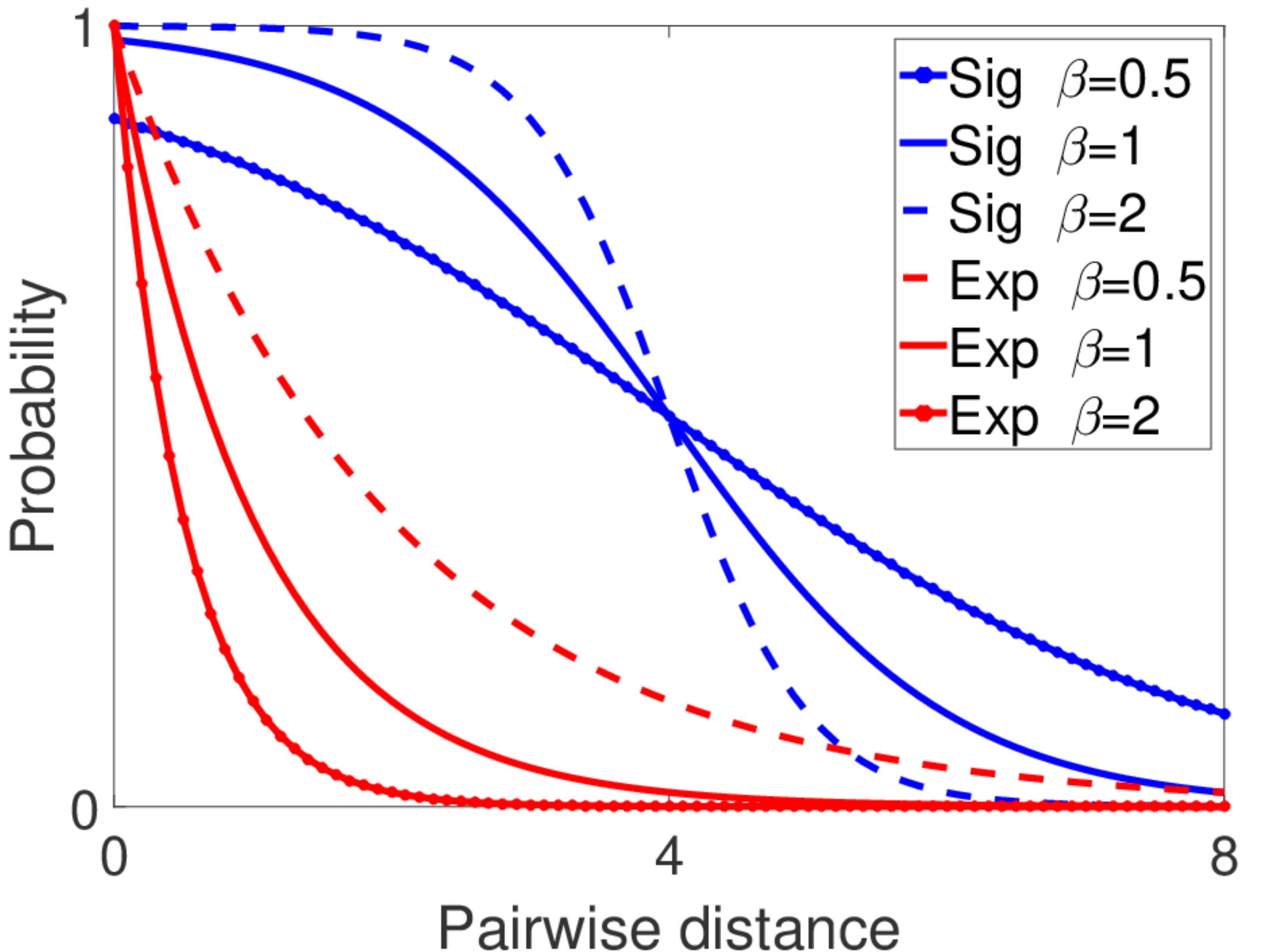}}
  \centerline{$(a)$ Probability}\medskip
\end{minipage}
\hfill
\begin{minipage}[b]{0.32\linewidth}
  \centering
  \centerline{\includegraphics[width=0.95\linewidth]{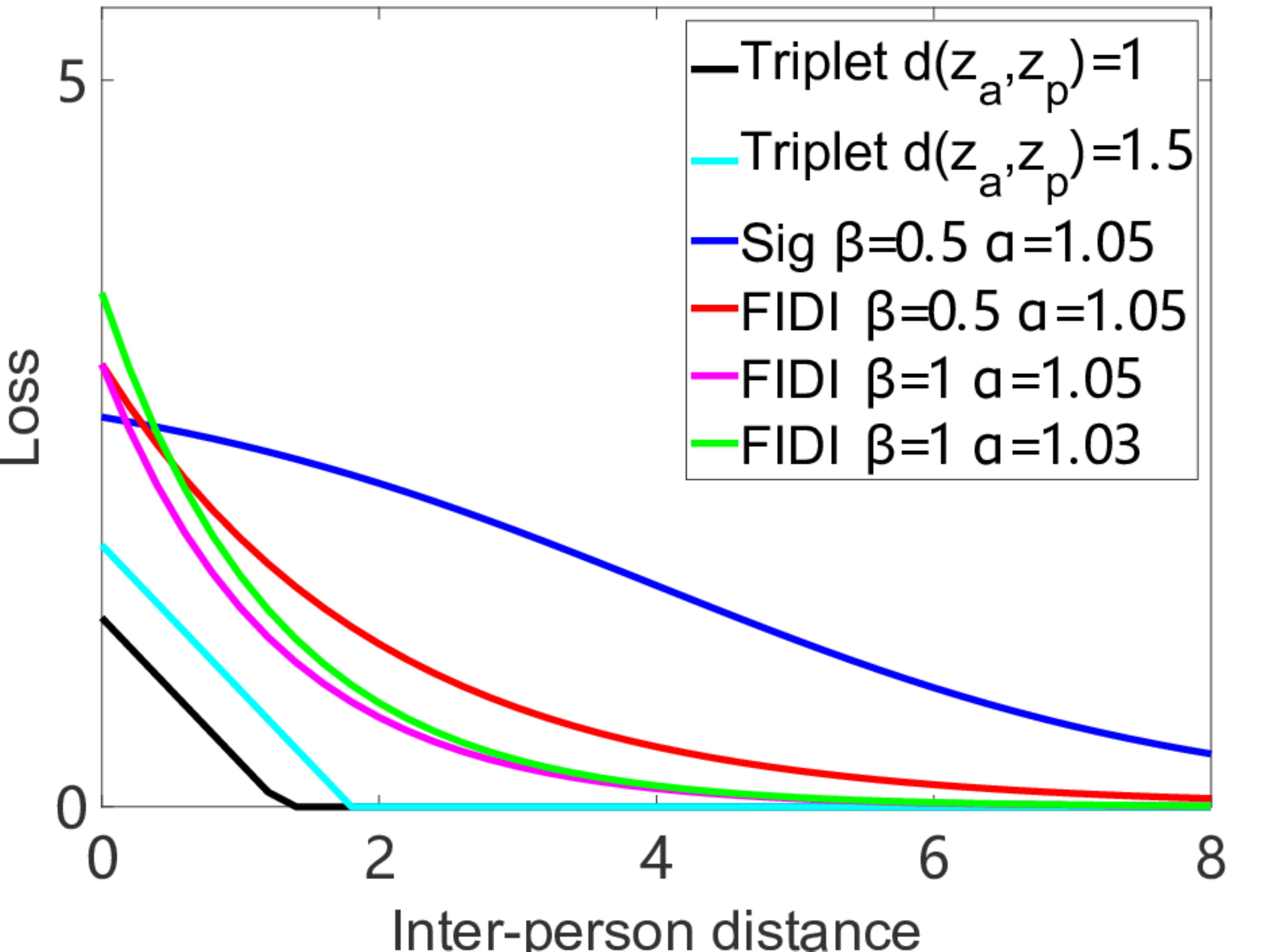}}
  \centerline{$(b)$ Loss ($k=0$)}\medskip
\end{minipage}
\hfill
\begin{minipage}[b]{0.32\linewidth}
  \centering
  \centerline{\includegraphics[width=0.95\linewidth]{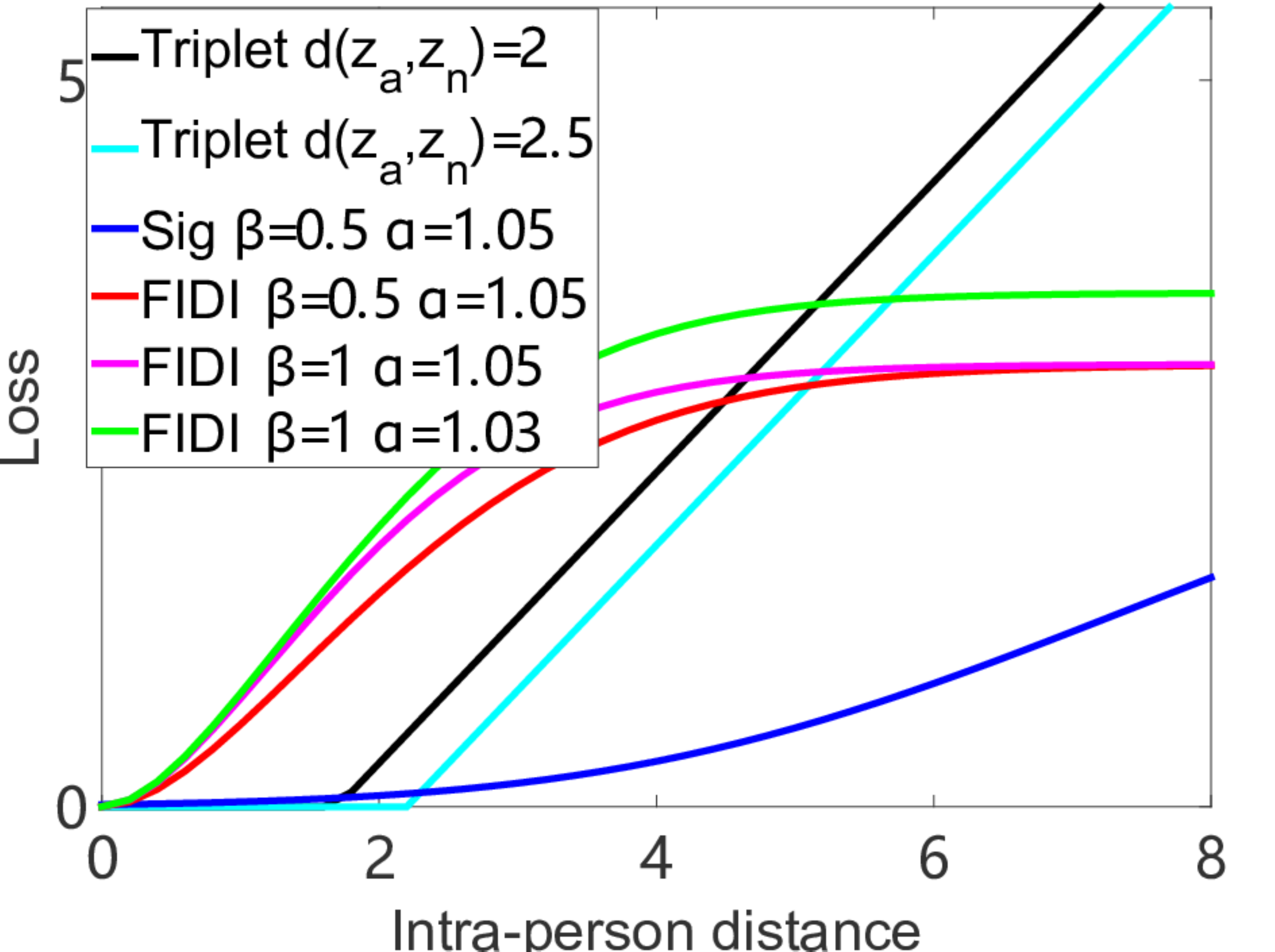}}
  \centerline{$(c)$ Loss ($k=1$)}\medskip
\end{minipage}
\caption{(a) Exponential vs. sigmoid distance-to-probability functions, (b) Loss w.r.t.\ inter-person distance and (c) Loss w.r.t.\  intra-person distance. One desired property of exponential distribution based distance-to-probability function is that its probability is exponentially sensitive to changes within a small distance. As shown in (b) and (c), our loss function exponentially punishes the similar/dissimilar pairs that have small distance whle imposing bounded loss to large distances.}
\label{fig:p_compare}
\end{figure*}

The procedure of our framework is as follows. It first uses a convolutional neural network to map image data into a low-dimensional space. Compared to quadruplet loss~\cite{chen2017beyond}, here this backbone network is not limited to any specific deep convolutional network structures. Then the proposed FIDI loss enforces a pairwise constraint to the projected features by applying exponentially increasing penalization to small differences and bounded loss to large differences. This enables the learner to adaptively capture the fine-grained differences while enforce a desired margin between the feature representations of different identities. Finally, we use a batch normalization layer and a fully connected layer without bias as the classifier, which is optimized using the cross entropy loss in Eqn.(\ref{cls}).

Particularly, the FIDI loss is built upon relative entropy~\cite{cover1991entropy}, a measure of the distance between two distributions. Let $\mathcal{K}$ be a known distribution of training image pairs, \textit{i.e.}, the ground truth identity labels, and $\mathcal{U}$ be an unknown distribution we aim to learn, then the FIDI loss is defined as follows:
\begin{equation}
\label{hap}
L_{fidi} = D(\mathcal{U}||\mathcal{K}) + D(\mathcal{K}||\mathcal{U}),
\end{equation}
where
\begin{equation}
\label{hap2}
\begin{split}
D(\mathcal{U}||\mathcal{K})= &\sum_{p_{ij} \in \mathcal{P}} u_{p_{ij}} \log \frac{\alpha u_{p_{ij}}}{(\alpha -1 ) u_{p_{ij}} + k_{p_{ij}}}, \\
\end{split}
\end{equation}
where $p_{ij}=\{\mathbf{x}_{i},\mathbf{x}_{j}\}$ is a pair of image samples and $\mathcal{P}$ is a collection of image pairs; $k_{p_{ij}}\in\mathcal{K}$ and $k_{p_{ij}}=1$ if the image pair $\mathbf{x}_{i}$ and $\mathbf{x}_{j}$ are from the same identity, and $k_{p_{ij}}=0$ otherwise; $u_{p_{ij}}$ is taken from an unknown distribution $\mathcal{U}$, which is the distribution of feature level relationship of image pairs in $\mathcal{P}$; and $\alpha > 1$ is a parameter to control the scale of $L_{fidi}$. Since $\mathcal{K}$ is the supervised information and is known \textit{a priori}, our target is to learn $u_{p_{ij}} \in \mathcal{U}$ such that the distribution $\mathcal{U}$ is close to $\mathcal{K}$ as much as possible.

The original relative entropy is one of the most popular and effective losses used in different learning tasks. However, it could not effectively reflect the true distance between distributions in our task due to the  asymmetric and the lack of fine-grained difference-aware characteristic. Our $L_{fidi}$ enhances it to achieve the following two main advantages:
\begin{itemize}
    \item Our loss is a symmetric metric with a desired inter-class margin.
    \item Our loss enforces fine-grained difference-aware penalization on small differences and bounded loss on large differences
\end{itemize}

\subsection{Exponential Loss on Images of Fine-grained Differences}

One key ingredient in Eqn.(\ref{hap}) is the distance-to-probability function $\eta$ that maps the distance in the representation space, $d(\mathbf{z}_{i},\mathbf{z}_{j})$, to the probability distribution $\mathcal{U}$, \textit{i.e.}, $u_{p_{ij}}=\eta(d(\mathbf{z}_{i},\mathbf{z}_{j}))$. In FIDI loss, we introduce an exponential distribution-based distance-to-probability function $\eta$ to effectively penalize hard samples. Particularly, $\eta$ is defined as follows:
\begin{equation}
\label{exponential}
u_{p_{ij}}= e^{-\beta d(\mathbf{z}_{i},\mathbf{z}_{j})},
\end{equation}
where $\beta$ is a parameter to control the scale of the probability distribution. We have $u_{p_{ij}} \rightarrow 0$ with increasing pairwise distance, and $u_{p_{ij}} \rightarrow 1$ in the opposite.

We use the exponential distribution-based $\eta$ because it is more sensitive and imposes more meaningful penalization on small differences compared to the commonly-used sigmoid function $\frac{1}{1+e^{-d}}$ or its advanced variant $\frac{1}{1+e^{-\beta d}}$~\cite{su2016deep,song2018mask,wang2018mancs}, where $d$ denotes the pairwise distance and the parameter $\beta$ controls the scale of the distribution shape.

Specifically, as shown in Figure~\ref{fig:p_compare}(a), the exponential distribution shape is significantly more sensitive to the distance than the sigmoid distribution shape, especially when the pairwise distance is small. As a result, as shown in Figure~\ref{fig:p_compare}(b-c), the exponential distribution based $\eta$ results in exponentially varying relative entropy loss w.r.t.\ both the intra- and inter-person distances, whereas the sigmoid distribution-based loss applies rather conservative penalization in such cases.

One main benefit brought by the exponentially sensitive penalization is the capability in learning the fine-grained difference of the image pairs. Specifically, as shown in Figure~\ref{fig:p_compare}(b), for image pairs that come from different persons but with small distances, the FIDI loss applies penalization inversely exponential to the distance and applies nearly zero loss to the pairs that have large inter-person distance; by contrast, the triplet loss may enforce no penalization on image pairs which have very small inter-person distance. In a similar sense, as shown in Figure~\ref{fig:p_compare}(c), for image pairs that come from the same person, no penalization is enforced by the triplet loss on the image pairs that have small intra-person distance; by contrast, the FIDI loss also applies exponential penalization to such cases.

The resulting FIDI loss-based model effectively learns fine-grained feature representations that are significantly improved over the triplet loss. The fine-grained difference-aware ability also enables the FIDI loss-based model to leverage the labeled data substantially more efficiently than its counterpart, resulting in more data-efficient learning.

\subsection{Bounded Loss on Images of Large Differences}

Unlike triplet loss that has an infinitely linearly increasing penalization w.r.t.\ images of large appearance differences, the FIDI loss has a bounded loss on the large differences, which effectively prevents the dominance of images of large differences in the optimization. Specifically, the bounded loss of the FIDI loss can be provided as follows.
\begin{equation}
\label{bound}
\begin{split}
& \lim_{u \rightarrow 0 } L_{fidi} = 0, \;\text{when} \  k=0;\\
& \lim_{u \rightarrow 0 } L_{fidi} = \log \frac{\alpha}{(\alpha -1 )},\; \text{when} \ k=1.\\
\end{split}
\end{equation}
This states that for image pairs from different identities, i.e., $k=0$, we have a lower loss bound of zero with $u$ approaching to zero. Recall that the pairwise distance increases as $u \rightarrow 0$. In other words, similar to the triplet loss, the FIDI loss does not penalize the image pairs if they are from different identities with a large distance in the new space. On the other hand, for image pairs from the same identity, while the triplet loss has an infinitely increasing loss, the FIDI loss imposes an upper loss bound of $\log \frac{\alpha}{(\alpha -1 )}$ w.r.t.\  increasing intra-person distance. This upper bound is controlled by the hyperparameter $\alpha$ and can be easily tuned during training.

As shown in Figure~\ref{fig:p_compare}(c), the punishment of triplet loss can be very large, given image pairs with very large intra-person distances. This hinders the triplet loss to learn the fine-grained differences from image pairs that have small intra-person distances. By contrast, the FIDI loss treats these samples equally and enforces a bounded penalization, preventing the domination of the large appearance differences over the counterpart small differences.

\subsection{Symmetric Metric with a Desired Margin}

Different from the original relative entropy that is asymmetric, our loss in Eqn.(\ref{hap}) is symmetric, as it is easy to see that we get the same results when switching $u_{p_{ij}}$ and $k_{p_{ij}}$. This characteristic eases the optimization of the feature learning and also helps learn more meaningful features.

Although the FIDI loss does not explicitly define a margin between intra- and inter-person image pairs as in the triplet loss, the FIDI loss can still achieve some implicit margins. This is because Eqn. (\ref{hap}) enforces the substantially small intra-person distances while at the same time encourages large inter-person distances, resulting in some implicit margins between intra- and inter-person image pairs. However, the margins are not directly predefined as in the triplet loss, but they are controlled by the parameter $\beta$ in Eqn. (\ref{exponential}).

\section{Experiments}
\label{sec:5}
\subsection{Datasets}
We evaluate the performance on four widely used person ReID datasets, including Market1501~\cite{zheng2015scalable}, DukeMTMC-ReID~\cite{zheng2017unlabeled}, CUHK03-D and CUHK03-L~\cite{li2014deepreid}, and two vehicle datasets, VeRi-776~\cite{liu2016eccv} and VehicleID~\cite{liu2016deep}.

\begin{figure}[htbp]
\centerline{\includegraphics[width=0.8\linewidth]{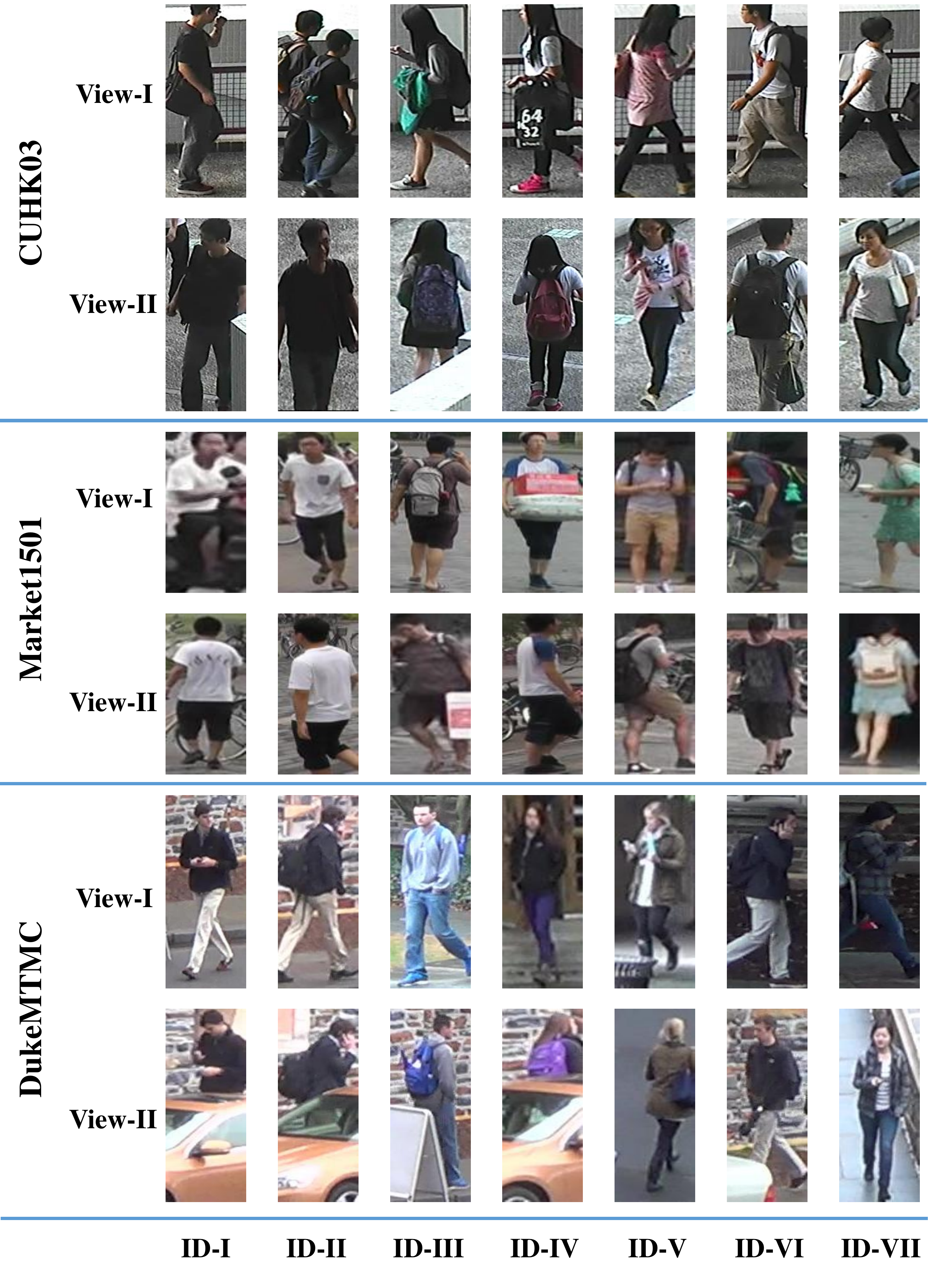}}
\caption{Images from three person reid datasets. We give two images from a same person with different views. There are many hard/easy examples from different/same person in these datasets. For example, the images of ID-I and ID-II in View-II look very similar. However, the images of same person of ID-IV and ID-VII in different views look very different. These datasets also contain many images with occlusion.}
\label{fig:dataset}
\end{figure}

\textbf{Market1501} is a large person ReID dataset containing 12,936 images from 751 identities in the training data, and 3,368 query images and 19,732 gallery images from 750 identities in the testing data. These images were captured from 6 different camera viewpoints with manual bounding boxes. There are about 17 images for each identity.

\textbf{DukeMTMC-ReID} is a subset of DukeMTMC~\cite{ristani2016performance} for person ReID. The images are cropped by hand-drawn bounding boxes. The data was taken from 8 cameras of 1,404 identities with respective 16,522, 2,228 and 17,661 images in the training, query and gallery sets.

\textbf{CUHK03-D} and \textbf{CUHK03-L} contain the same image set with 14,096 images from 1,467 identities captured from two cameras in CUHK campus, but their identity-bounding box were created by different methods. CUHK03-D used pedestrian detectors to create the bounding boxes while that of CUHK03-L was manually labeled. The pedestrian detector-based method is more challenging than the manually labeled one since the former is less accurate.

\textbf{VeRi-776} is a vehicle dataset in which all the images were captured in natural and unconstrained traffic environment. It contains about 50,000 images of 776 vehicles across 20 surveillance cameras with different orientations. This dataset is widely used in vehicles re-identification tasks because  each image is captured from 2 to 18 viewpoints with different illuminations and resolutions. These images are also labeled with bounding boxes over the whole vehicle body.

\textbf{VehicleID} is a large-scale vehicle dataset that contains 221,763 images with 26,267 vehicles. All the images were captured from multiple surveillance cameras with no overlapping. There are three test subsets with different sizes and we use the largest test set which contains 20,038 images of 2,400 vehicles.

Note that the person/vehicle identities in the training and testing sets have no overlapping in all the used datasets. An image example is given in Figure.~\ref{fig:dataset}, in which we give two images from the same person with different views. There are many hard/easy examples from different/same person in these datasets. For example, the images of ID-I and ID-II in View-II look very similar and the images of same person of ID-IV and ID-VII in different views look very different. These datasets also contain many images with occlusion.

\subsection{Evaluation Protocol} Following the standard protocol in~\cite{song2018mask,wang2018mancs,sun2018pcb,chen2018person,sun2019perceive}, we use Cumulated Matching Characteristics (CMC) and  mean average precision (mAP) to evaluate the performance on all datasets. We report the cumulated matching accuracy at rank 1 (R-1 for short) and the mAP value of the retrieval performance. Specifically, for all queries, we compute

\begin{equation}
R1=\sum_{q=1}^Q r_{1} / Q,
\end{equation}
where $Q$ is the number of queries and $r_{1}$ is defined as

\begin{equation}
\label{r1}
r1=
\begin{cases}
1, & \text{ the first top-ranked sample is the query identity}\\
0, & \text{ otherwise},
\end{cases}
\end{equation}

The mean average precision (mAP) is defined as

\begin{equation}
\label{mAP}
mAP = \sum_{q=1}^{Q} AveP(q) / Q,
\end{equation}
where $AveP(q)$ is the average precision (AP) for a given query $q$.

Note that all the reported results here do not involve re-ranking which may be used as an extra step to further improve the accuracy.

\subsection{Understanding the Resulting Feature Representations}
We aim to understand the effectiveness of feature representations by looking into the fidelity and the saliency map of learned features.
\subsubsection{Fidelity of Feature Representations}
The feature representations fidelity  measures how faithful the obtained feature represents the expectation, \textit{i.e.} intra-person distances should be larger than inter-person distances. To efficiently evaluate this type of fidelity, we consider the number of erroneous cases where (i) anchor images have smaller inter-person distances than the their maximal intra-person distances, termed Error-I, or (ii) anchor images have larger intra-person distances than their minimal inter-person distances, termed Error-II. We count these two types of erroneous cases using feature representations of four different methods, including pre-trained features extracted from a pre-trained ResNet50\footnote{https://github.com/kaiminghe/deep-residual-networks} (PF) and features obtained by fine-tuning ResNet50 using respectively batch-hard constrastive loss (BCL), batch-hard triplet loss (BTL) and our proposed loss (FIDI). The statistics of erroneous cases on three person ReID benchmarks are reported in Table~\ref{tab:number_of_hard_examples}.

\begin{table*}[htbp]
\caption{Average erroneous distance cases over all images of each dataset in four feature spaces. Error-I refers to the average number of anchor images which have smaller inter-person distances than their maximal intra-person distances, while Error-II is the average number of anchor images which have larger intra-person distances than their minimal inter-person distances. PF refers to Pre-trained Features extracted from a pre-trained ResNet50. BCL, BTL and FIDI are feature spaces resulted by fine-tuning ResNet50 using respectively batch-hard constrastive loss, batch-hard triplet loss and our proposed loss. The average number of images per identity is 9.6 in CUHK03, 17 in Market1501 and 23 in DukeMTMC. The best results are boldfaced in each group.}
\centering
\scalebox{1.1}{
\begin{tabular}{c|c|c c|c c|c c}

\hline
\hline
\multirow{3}{*}{\textbf{Data}}& \multirow{3}{*}{\textbf{Method}}& \multicolumn{2}{c|}{\textbf{CUHK03}}  & \multicolumn{2}{ c| }{\textbf{Market1501}} & \multicolumn{2}{ c }{\textbf{DukeMTMC}} \\
\cline{3-8}
& & \multicolumn{2}{c|}{9.6 images/ID} & \multicolumn{2}{c|}{17 images/ID} & \multicolumn{2}{c}{23 images/ID} \\
\cline{3-8}
& & Error-I & Error-II & Error-I& Error-II & Error-I& Error-II \\
\hline

\multirow{4}{*}{\textbf{Training Data}}& PF & 4316& 7 & 9596 & 15 & 13677 & 19 \\
& BCL &0.008 & 0.010 & 1.977& 0.397 & 6.360& 4.420 \\
& BTL & \textbf{0.005} & \textbf{0.008} & \textbf{1.905}& 0.295 &\textbf{3.310} & 3.655 \\
& FIDI &0.252 & 0.023 &2.072 &\textbf{0.261}  &3.721 & \textbf{1.807} \\
\hline
\multirow{4}{*}{\textbf{Testing Data}}& PF & 2973 & 7.617 & 11687 & 506.7 & 13258 & 24.15 \\
& BCL & 85.06& 6.600 & 276.3 & 19.18 & 925.7& 20.89 \\
& BTL & 95.11 & 6.611 & 261.5 & 18.36 & 910.6 & 20.61 \\
& FIDI &\textbf{45.02} & \textbf{6.310} &\textbf{229.1} &\textbf{16.50}  &\textbf{819.3} & \textbf{19.54} \\
\hline
\hline
\end{tabular}
\label{tab:number_of_hard_examples}
}
\end{table*}

It is clear from Table~\ref{tab:number_of_hard_examples} that pre-trained features would result in a large number of erroneous cases, especially the Error-I cases. This indicates that most images of difference identities exhibit large similar appearance in both training and testing data, leading to small inter-person distances. The datasets also contain some Error-II cases that may be seen as outliers, because intra-person distances is rarely larger than minimal inter-person distances. To address these issues, models should be able to effectively learn the small appearance differences while prevent the impact of the outlying cases. After fine-tuning the models using either BCL, BTL or FIDI, the number of erroneous cases is significantly reduced in both training and testing data. In training data, BCL and BTL perform very well in enforcing intra-person distances to be smaller than inter-person distances, often achieving smaller error rates than FIDI. However, they perform significantly less effective than FIDI in the testing data, especially on the Error-I measure. This may indicate that both BCL and BTL overfit the training data rather than capturing the fine-grained appearance differences to distinguish the inter-person images. By contrast, with exponentially large penalization, FIDI enforces the models to learn any possible fine-grained appearance differences in the training data. Since the fine-grained differences are typically very difficult to learn, for some cases, even for humans, FIDI does not perform as well as BCL and BTL in the training data. However, its capability of discriminating the fine-grained differences pays off in the testing data.

\subsubsection{Attention Maps}

\begin{figure}[htbp]
\centerline{\includegraphics[width=0.8\linewidth]{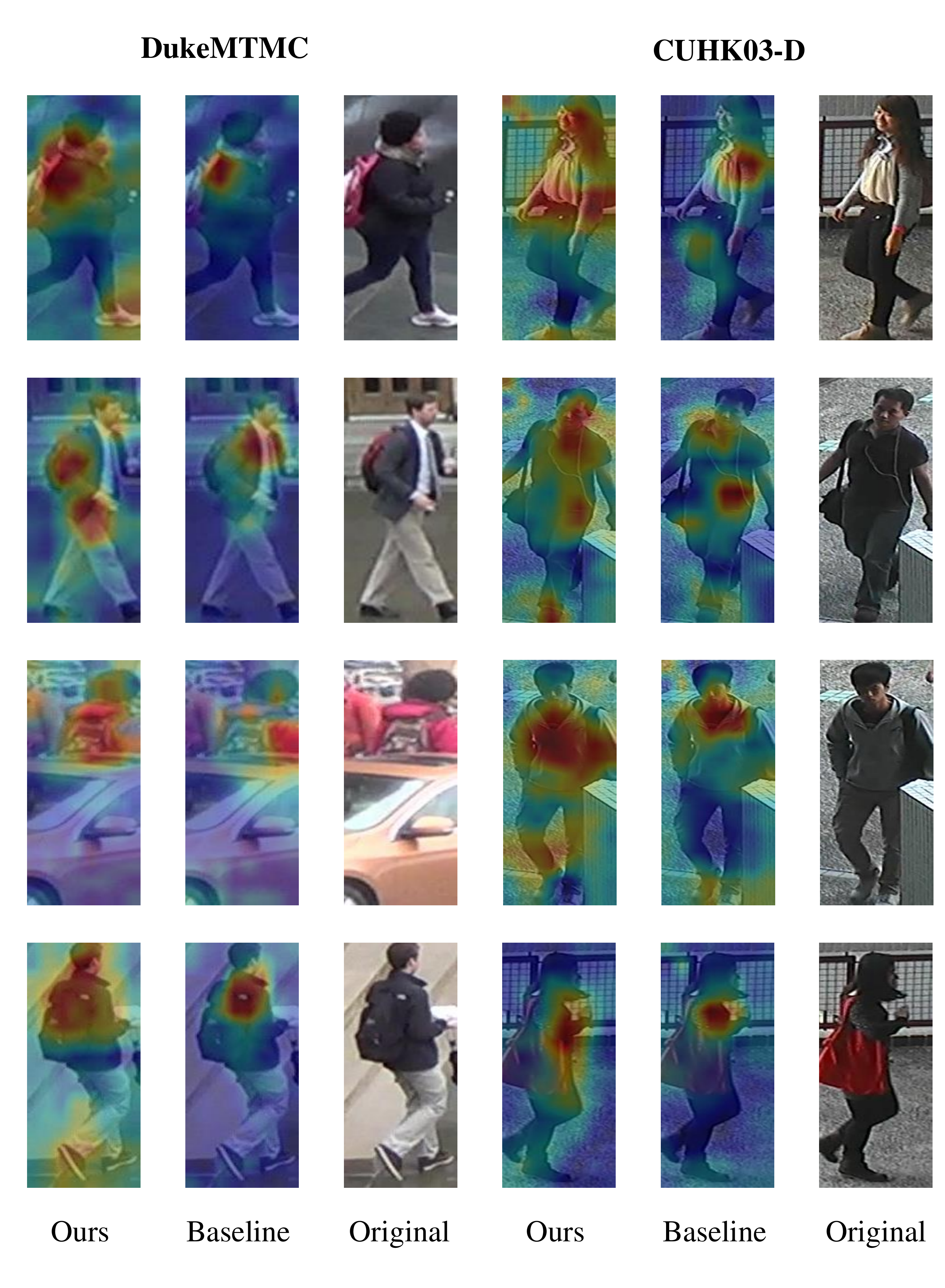}}
\caption{Visualization of attention maps of our FIDI loss-enabled model (Ours) and the batch-hard triplet loss-enabled model (Baseline). Our method learns diverse important attention, but Baseline only focuses on small discriminative parts. The diverse attention maps from Ours span over the whole person rather than some local areas in Baseline.}
\label{fig:vis}
\end{figure}

We further examine the resulting attention maps of our loss and the competing loss functions. We focus on comparing our FIDI loss to the BTL loss, because BTL is generally more effective and is much more widely-used than BCL in person ReID. Specifically, these two losses are plugged into one of the best ReID models, Baseline~\cite{luo2019bag}. The attention maps are then obtained by applying the Grad-CAM visualization method~\cite{selvaraju2017grad} with Baseline to create pix-wise gradient visualizations. The attention maps on the last output feature maps are shown on Figure~\ref{fig:vis}. The BTL-enabled Baseline highlights single discriminative parts only, which may correspond to the parts that have large appearance differences to other images. In contrast, our FIDI loss-enabled Baseline can effectively attend to diverse large and small discriminative parts in different cases, \textit{e.g.}, shoes and heads in identity images taken different angles, different accessories and occluded identities. For example, in the 1st row in Figure~\ref{fig:vis}, despite different angles and identities, our method can consistently pay attention to both small (shoes and heads) and large (the main body dress) discriminate parts, while the competing method focuses on a small discriminative region of the main body only. This demonstrates that the BTL loss-based models can be dominated and swayed by large appearance differences.  Therefore, their attention is normally on single highly discriminative parts. In contrast, our loss can effectively drive the ReID models to pay attention on different body parts by enforcing the importance of distinguishing fine-grained differences.

\subsubsection{Summary of Comparison}
Overall, by enforcing exponentially large penalization on images of small appearance differences and bounded penalization on images of large differences, our FIDI pairwise loss brings in two major benefits compared to existing widely-used pairwise and triplet losses. First, the FIDI-enabled models can effectively capture fined-grained appearance differences, where the competing methods fail. This significantly improves the feature representations as demonstrated by significantly small errors in testing data in Table~\ref{tab:number_of_hard_examples}. Second, as illustrated in Figure~\ref{fig:vis}, our loss effectively pushes the ReID models to attend to diverse discriminative parts since fine-grained differences may appear in different body parts. This is important for distinguishing different identities with some similar appearances,\textit{ e.g.}, in dress, shoes and/or accessories. Models embodied with our loss would enjoy above two factors that are critical to accurate person ReID.

\subsection{Enabling Different Type of Person ReID Models in Real-world Datasets}

To have a comprehensive evaluation on real-world datasets, the FIDI loss is used to replace the batch-hard triplet loss in three types of recent state-of-the-art approaches, including data augmentation, global feature and striping approaches. Specifically, we choose the best performer(s) in each type of these approaches and them simply replace the triplet loss with our proposed FIDI loss, with all the other modules unchanged. The batch size and the number of identities in each batch are respectively set to 128 and 8 by default. The hyperparameters $\alpha$ and $\beta$ in the FIDI loss are tuned via cross validation for each data set.

\subsubsection{Enabling Data Augmentation Methods}

This section compares our loss to several data augmentation-based methods, including GAN-based methods~\cite{qian2018pose,zhong2018camstyle} and segmentation-based masking methods~\cite{kalayeh2018human,sun2017svdnet}. Note that, these methods employ other networks to generate images to obtain semantic information, which brings extra computational consumption. Baseline1~\cite{luo2019bag} without data augmentation, \textit{i.e.}, random erasing, is the best performer. Therefore we plugged the FIDI loss into this method. Note that Baseline1 contains a center loss and we discard this loss in our Baseline1 model by replacing the triplet loss with our FIDI.

\renewcommand{\arraystretch}{1}
\begin{table*}[htbp]
\centering
\caption{MAP and R-1 of different methods on four benchmark datasets. BCL and BTL respectively denote batch contrastive loss and batch triplet loss. The best performance per group is boldfaced.}
\scalebox{1.1}{
\begin{tabular}{c|l|c c|c c|c c|c c}
\hline
\multirow{2}{*}{\textbf{Type}} & \multirow{2}{*}{\textbf{Methods}} &  \multicolumn{2}{ |c| }{\textbf{Market1501}} & \multicolumn{2}{ |c }{\textbf{DukeMTMC}}& \multicolumn{2}{ |c }{\textbf{CUHK03-D}}& \multicolumn{2}{ |c}{\textbf{CUHK03-L}}\\
\cline{3-10}
& & mAP & R-1 & mAP & R-1 & mAP & R-1 & mAP & R-1 \\
\hline
\multirow{6}*{\shortstack{Data\\Augumentation}}
& SPReID~\cite{kalayeh2018human}   & 81.3 & 92.5 & 71.0 & 84.4 & -&-&-&-\\
& Camstyle~\cite{zhong2018camstyle}  & 68.7 & 88.1 & 53.5 & 75.3 &  - & - & - & -\\
& PN-GAN~\cite{qian2018pose}  & 72.6 & 89.4 & 53.2 & 73.6 & - & - & - & -\\
& SVDNet~\cite{sun2017svdnet}  &  62.1 & 82.3 & 56.8 & 76.7 & 37.3 & 41.5 & 37.8 & 40.9 \\
& Baseline1 (BTL)~\cite{luo2019bag} & 82.3 & 93.5 & 71.0 & 84.9 & 52.5 & 54.2 & 55.3 & 56.3  \\
& Baseline1 (FIDI)  &  \textbf{84.5} & \textbf{94.2} & \textbf{73.4} & \textbf{86.0} & \textbf{63.5} & \textbf{66.1} & \textbf{67.1} & \textbf{69.1} \\\hline
\multirow{9}*{\shortstack{Global\\ Feature}}
& TriNet~\cite{hermans2017defense}  & 69.1& 84.9 & - & - & -& -& -& - \\
& AWTL~\cite{ristani2018features}  & 75.7 & 89.5 & 63.4 & 79.8 & - & - & - & -\\
& AOS~\cite{huang2018adversarially}  & 70.4  & 86.4  & 62.1  & 79.1 & 43.3  & 47.1 & -  & - \\
& GSRW~\cite{shen2018deep} & 82.5  & 92.7  & 66.4  & 80.7 &  - & - & -  & - \\
& Mancs~\cite{wang2018mancs}  & 82.3  & 93.1  & 71.8  & 84.9 & 60.5 & 65.5 & 63.9 & 69.0 \\
& BCL~\cite{zhang2018discriminative} & 67.6  & 86.4  & 58.6  & 78.2 &- &- & -&-\\
& CL \cite{sun2020circle}  & 84.9  & 94.2  & -  & - &  -&  -& - & - \\
& Baseline2 (BTL)\cite{luo2019bag}  & 85.9  & \textbf{94.5}  & 76.4  & 86.4 & 58.2 & 60.5 & 60.2 & 62.1 \\
& Baseline2 (BCL)  & 84.0  & 92.3  & 74.5  & 83.6  & 60.5 & 63.3 & 64.9 & 66.4\\
& Baseline2 (FIDI)  & \textbf{86.8} & \textbf{94.5} & \textbf{77.5} &\textbf{88.1} & \textbf{69.1} & \textbf{72.1} & \textbf{73.2} & \textbf{75.0} \\
\hline
\multirow{10}*{\shortstack{Striping}}
& AlignedReID~\cite{zhang2017alignedreid}  & 77.7 & 90.6 & 67.4 & 81.2 &  - & - & - & -\\
& MLFN~\cite{chang2018multi}  & 70.4  & 86.4  & 62.1  & 79.1 & 47.8  & 52.8 & 49.2  & 54.7 \\
& PCB~\cite{sun2018pcb}  & 77.4  & 92.3  & 65.3  & 81.9 &  53.2 & 59.7 & -  & - \\
& IANet~\cite{hou2019interaction}  & 83.1  & 94.4  & 73.4  & 87.1 &  - & - & -  & - \\
& PL-Net~\cite{yao2019deep} & 69.3  & 88.2  & -  & - &  - & - & -  & - \\
& MCG~\cite{zhai2019defense}  & 78.3  & 92.6  & 69.4  & 84.7 &  - & - & 55.3  & 61.7 \\
& BDB~\cite{dai2019batch}  & 84.3  & 94.2  & 72.1  & 86.8 &  69.3 & 72.8 & 71.7  & 73.6 \\
& BDB (FIDI)  & 85.2  & 94.8  & 74.5  & 88.6 &  71.7 & 74.5 & 73.8  & 76.9 \\
& MGN~\cite{wang2018learning}  & \textbf{86.9} & \textbf{95.7} & 78.4 & 88.7 & 66.0 & 66.8 & 67.4 & 68.0 \\
& MGN (FIDI) & \textbf{86.9}  & 95.4  & \textbf{79.8}  & \textbf{89.7} &  \textbf{73.0} & \textbf{76.1} & \textbf{76.3}  & \textbf{78.9} \\
\hline

\end{tabular}
\label{tab:MAP}
}
\end{table*}

The comparison results are shown in the second row in Table~\ref{tab:MAP}. Although Baseline1 has significantly outperformed all the other competing methods in this category, the FIDI loss-enabled Baseline1 can still consistently beat the original Baseline1 in both mAP and R-1 across all the four datasets. Particularly, the improvement is significantly larger on the challenging datasets than the relatively easy ones, \textit{ e.g.}, the improvement can be as large as 20.9\%-21.3\% in mAP and 22.1\%-22.7\% in R-1 on the two CUHK03 datasets whereas it is 2.7\%-3.4\% in mAP and 0.7\%-1.3\% in R-1 on Market1501/DukeMTMC. This demonstrates that the FIDI loss-enabled Baseline1 not only inherits the superior capability as in the original Baseline1 but also leverages the fine-grained difference-aware ability of the FIDI loss to learn extra discriminative information from the hard samples. This is especially true for the two challenging CUHK03 datasets in which we have much less images and the triplet loss-based models become overfitting (see Table~\ref{tab:number_of_hard_examples} for detail).

\subsubsection{Enabling Global Feature-based Methods}
We then examine the plugging of the FIDI loss into the global feature-based methods. There are seven methods for comparison, including TriNet~\cite{hermans2017defense}, AWTL~\cite{ristani2018features}, AOS~\cite{huang2018adversarially}, GSRW~\cite{shen2018deep}, Mancs~\cite{wang2018mancs}, BCL~\cite{zhang2018discriminative} and Baseline2~\cite{luo2019bag}. These methods only employ simple single branch structure for training, which have less parameters to learn. All methods have only one pipeline with the basic ResNet50 as the backbone and use the feature representations obtained after global pooling. Note that, Baseline2~\cite{luo2019bag} is the Baseline1 with random erasing data augmentation. We also discard the centre loss and replace the triplet loss by our FIDI loss. We not only report the results of BCL~\cite{zhang2018discriminative} but also the results of Baseline2 (BCL), which is the Baseline2~\cite{luo2019bag} with BTL function being replaced by the BCL function from ~\cite{zhang2018discriminative}.

The results are given in the third row in Table~\ref{tab:MAP}. It is clear that the FIDI loss-enabled Baseline2 consistently enhances the best performer in this group of methods, the original Baseline2, with significant improvement on the two CUHK03 datasets by 18.8\%-20.3\% in mAP and 19.2\%-20.9\% in R-1. This is because the FIDI loss-enabled Baseline2 can still gain the full benefits brought by a bag of different tricks used in Baseline2 while at the same time significantly improving Baseline2 when the datasets become more challenging.

\subsubsection{Enabling Striping-based Methods}\label{subssec:striping}

Lastly the FIDI loss is evaluated with the stripe-based methods, including some recent promising methods BDB~\cite{dai2019batch} and MGN~\cite{wang2018learning}\footnote{DSA~\cite{zhang2019densely} also achieves state-of-the-art results on CUHK03-D and CUHK03-L datasets, but we cannot plug our loss into it since its source code is not available.}. These methods are typically much more difficult to train and are computationally expensive than the other methods, because they involve multi-branch complex network structures. Since there is no consistent superiority of MGN and BDB over each other, we plug our FIDI loss into both methods.

The results are shown in the last row in Table~\ref{tab:MAP}. We can see that the performance of both BDB and MGN is substantially improved in nearly all cases on the four datasets. Particularly, the FIDI loss consistently enhances BDB in both mAP and R-1 across all cases, especially lifting its mAP performance by 1.1\% on Market1501, 4.2\% on DukeMTM, 3.4\% on CUHK03-D and 3.7\% on CUHK03-L. The FIDI loss significantly improves MGN by 10.1\%-13.2\% in mAP and 13.9\%-16.0\% in R-1 on the two complex CUHK03 datasets. The FIDI loss-enabled MGN only works comparably well to, or less effectively than, the original MGN on Market1501. This may be due to that Market1501 is a simple and small dataset while MGN is a model with very complex architecture. Therefore, training MGN with our FIDI loss may lead to overfitting on this dataset.

\begin{figure*}[htbp]
\centerline{\includegraphics[width=0.95\linewidth]{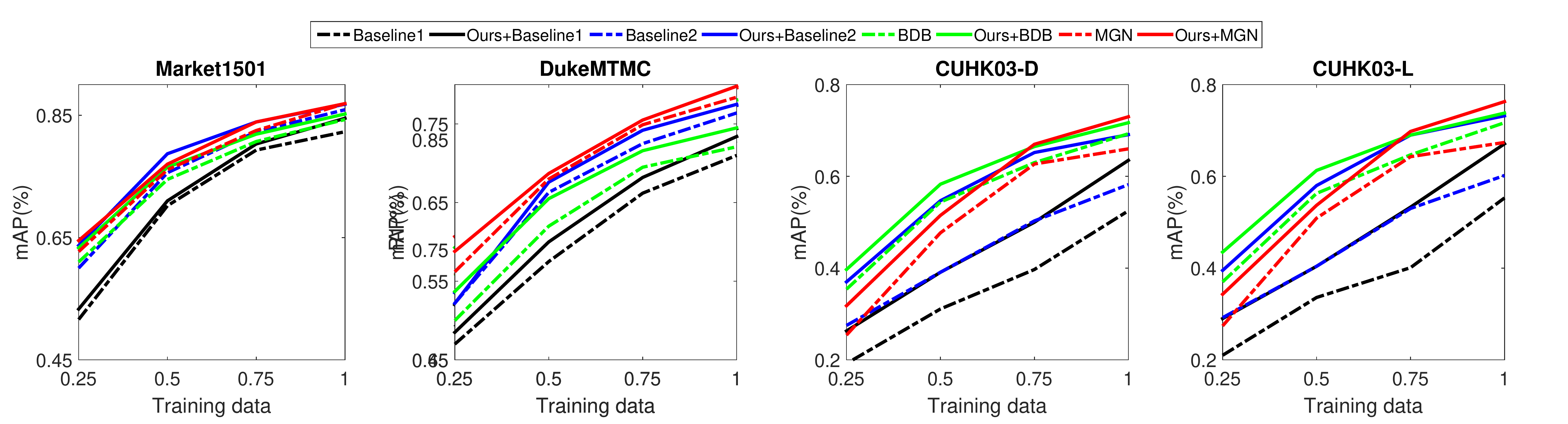}}
\caption{mAP results on four datasets with varying percentage of training data.}

\label{fig:reduced_map}
\end{figure*}

\begin{figure*}[htbp]
\centerline{\includegraphics[width=0.95\linewidth]{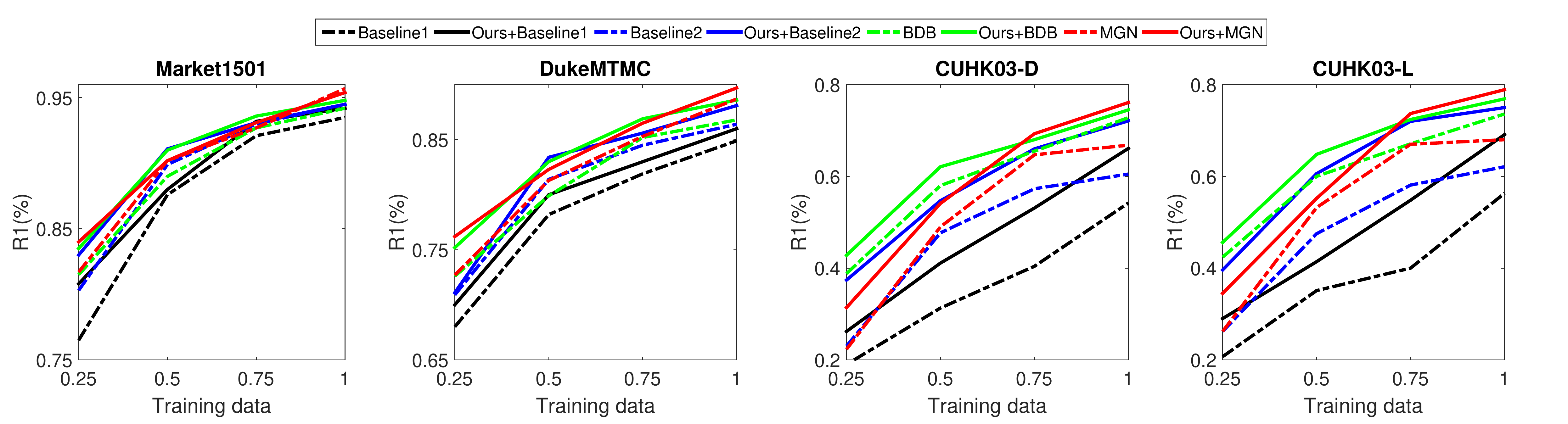}}
\caption{R1 results on four datasets with varying percentage of training data.}

\label{fig:reduced_r1}
\end{figure*}

\subsubsection{Summary of Comparison}
Overall, three main observations can be drawn across all the comparisons in Table~\ref{tab:MAP}. First, our FIDI loss consistently and substantially improves all three types of recently proposed triplet loss-based state-of-the-art methods by a large margin on DukeMTMC and the two CUHK03 datasets. It is especially true on the complex CUHK03 datasets where the plugin of the FIDI loss typically results in 10\%-20\% improvement in both mAP and R-1, but the FIDI loss may not have clear advantages over the triplet loss on simple and/or small datasets such as Market1501. Second, by using the FIDI loss, simple models can perform substantially better than the complex models that use the triplet loss, e.g.,  `Baseline2 (FIDI)' vs. BDB on Market1051, DukeMTMC and CUHK03-L. Third, the superiority of the FIDI loss sets new state-of-the-art results on DukeMTM, CUHK03-D and CUHK03-L, achieving 3.1\%-7.2\% improvement in mAP and 2.3\%-4.6\% improvement in R-1 over the prior best performance on the last two datasets.

\subsection{Enhancing Data Efficiency}

This section evaluates the data efficiency of the FIDI loss-enabled models. To do this, we reduce the training data by randomly removing 25\% identities each step. The mAP results are given in Figure~\ref{fig:reduced_map} and Figure~\ref{fig:reduced_r1}.

It is clear that the FIDI loss-enabled models outperform their corresponding counterparts in all the training data settings across the four datasets, with substantial improvement in most cases. The performance of the proposed loss on the easy datasets Market1501 and DukeMTMC is mainly due to its shared key similar properties as the triplet loss, e.g., having an inter-class margin, while the superiority of our loss on the challenging datasets CUHK03-D and CUHK03-L is due to its fine-grained difference-aware capability and the bounded loss for easy samples. It is very impressive that even when three FIDI loss-enabled models use 25\% less training data, they still can perform substantially better than the same models that use the triplet loss by a margin of at least 7.3\% on CUHK03-D and CUHK03-L. This indicates that when handling challenging data, using a fine-grained difference-aware loss function is a much more cost-effective way than increasing the training data.

\subsection{Beyond Person ReID: Enabling Vehicle ReID}
To further evaluate the capability of our proposed loss, we evaluate the performance of the Baseline2 (FIDI) on two vehicle ReID datasets, VeRi-776~\cite{liu2016eccv} and VehicleID~\cite{liu2016deep}.

\subsubsection{Comparison with State-of-the-art Vehicle ReID Methods}

We compare our method with 11 state-of-the-art vehicle ReID methods, including S-CNN~\cite{shen2017learning}, AAVER~\cite{khorramshahi2019dual}, VAMI ~\cite{zhou2018aware}, PROVID~\cite{liu2017provid}, MSVR ~\cite{kanaci2018vehicle}, FDA-NET~\cite{lou2019veri}, OIFE~\cite{wang2017orientation}, RAM~\cite{liu2018ram}, FACT~\cite{liu2016eccv}, P-R~\cite{he2019part} and Baseline2~\cite{luo2019bag}. The P-R and Baseline2 are more recent methods that have better performance than others.

The results are shown on Table~\ref{tab:MAP_Ve}. We can see from the results that the Baseline2 (FIDI) outperforms most vehicle ReID methods by a large margin. Compared to Baseline2, our method achieves 1.3\% - 2.6\% improvement on mAP and and 0.6\% - 0.7\% improvement on R-1. This demonstrates that the proposed FIDI loss can effectively generalize from person ReID to vehicle ReID.

\renewcommand{\arraystretch}{1.05}
\begin{table}[htbp]
\centering
\caption{MAP and R-1 performance on vehicle ReID datasets.}
\scalebox{1.1}{
\begin{tabular}{c|c c|c c}
\hline
\hline
\multirow{2}{*}{\textbf{Methods}}& \multicolumn{2}{ c| }{\textbf{VeRi-776}} & \multicolumn{2}{ c }{\textbf{VehicleID}}\\
& mAP & R-1 & R-1 & R-5 \\
\hline
S-CNN~\cite{shen2017learning}  & 58.3  & 83.5 & - & - \\
AAVER~\cite{khorramshahi2019dual} & 66.4& 90.2& 63.5& 85.6   \\
VAMI ~\cite{zhou2018aware}  & 50.1  & 77.0 & - & - \\
PROVID~\cite{liu2017provid}  & 53.4  & 81.6 & - & - \\
MSVR ~\cite{kanaci2018vehicle}  & 49.3  & 88.6 & 63.0 & 73.1   \\
FDA-NET~\cite{lou2019veri} & 55.5& 84.3& 55.5& 74.7   \\
OIFE~\cite{wang2017orientation}  & 51.4  & 92.4 & 67.0 & 82.9   \\
RAM~\cite{liu2018ram} & 61.5 & 88.6 & 67.7 & 84.5   \\
FACT~\cite{liu2016eccv}   & 27.8  & 61.4 & - & - \\
P-R~\cite{he2019part}  & 74.3  & 94.3 & 74.2  & 86.4   \\
Baseline2~\cite{luo2019bag}  & 75.7  & 95.2 & 77.5  & 91.0   \\
Baseline2 (FIDI)   & \textbf{77.6}  & \textbf{95.7} & \textbf{78.5}  & \textbf{91.9}   \\

\hline
\hline
\end{tabular}
}
\label{tab:MAP_Ve}
\end{table}

\subsubsection{Visualization of Ranking Results}
\begin{figure}[htbp]
\begin{minipage}[b]{1\linewidth}
  \centering
  \centerline{\includegraphics[width=1\linewidth]{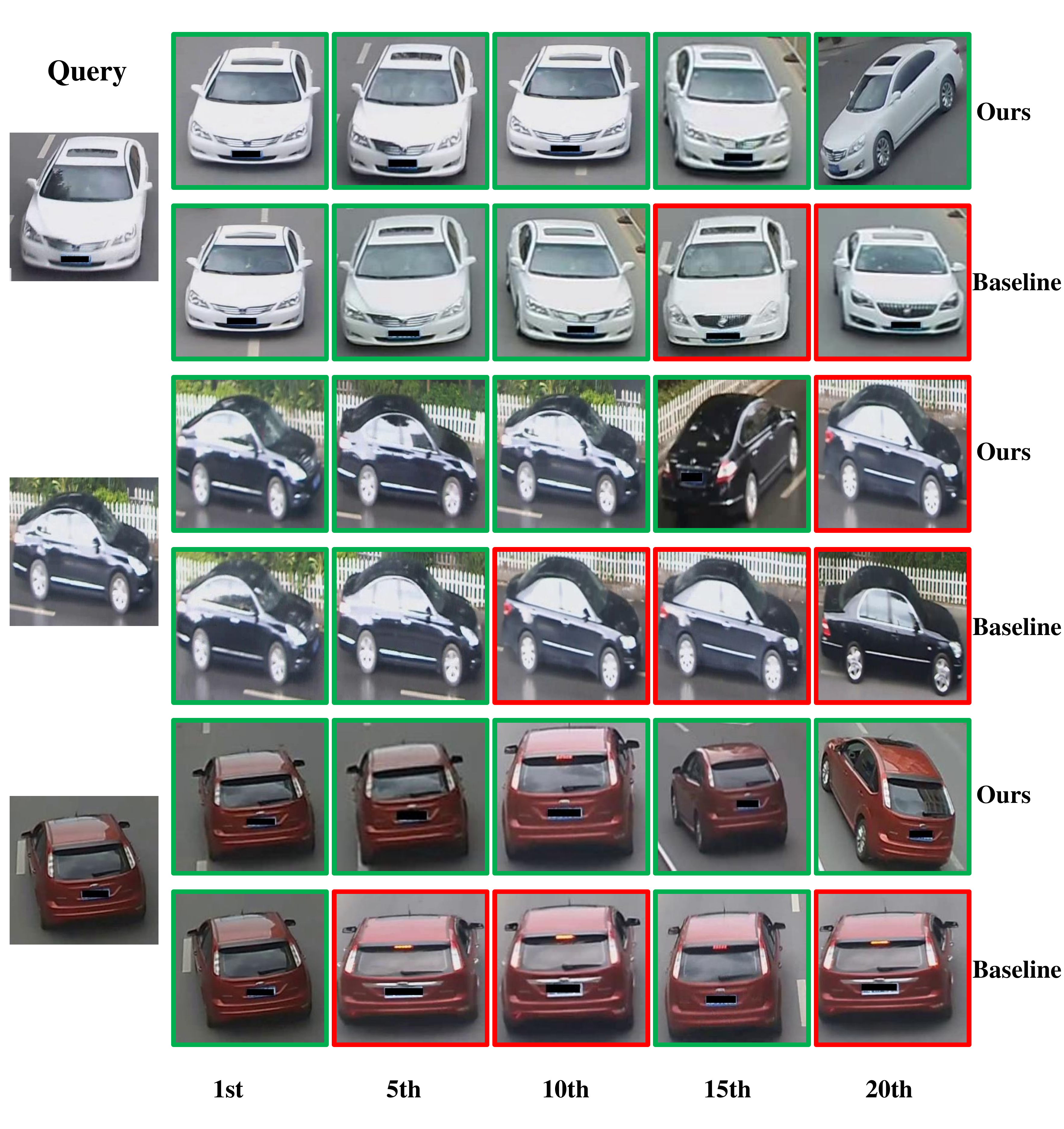}}
\end{minipage}
\caption{Exemplar Ranking Results from Our Model, Baseline2 (FIDI), and the Original Baseline2.}
\label{fig:vis_ve}
\end{figure}

To provide a more straightforward illustration of the effectiveness, we present a set of visual image ranking results for vehicle ReID on VeRi-766 in Figure~\ref{fig:vis_ve}. We only show the results of Baseline2 (FIDI) and Baseline2~\cite{luo2019bag} because Baseline2 has better performance than all the other competing methods. As shown in Figure~\ref{fig:vis_ve}, with the increase of returned images, the accuracy of Baseline2 decreases. For example, the 15-th and 20-th returned images of Baseline2 in the 2nd row are highly similar to the query image but they are actually different vehicles. This happens because Baseline2 fails to distinguish the fine grained appearance differences in the front of the vehicles. In contrast, Baseline2 (FIDI) can effectively capture the fined-grained differences and thus is able to return the images of the same vehicles taken from different viewpoints rather than the vehicles that are different from the query vehicle but share large similar appearance.

\subsection{Implication for Parameter Tuning}

\begin{figure}[htbp]
\begin{minipage}[b]{1\linewidth}
  \centering
  \centerline{\includegraphics[width=1\linewidth]{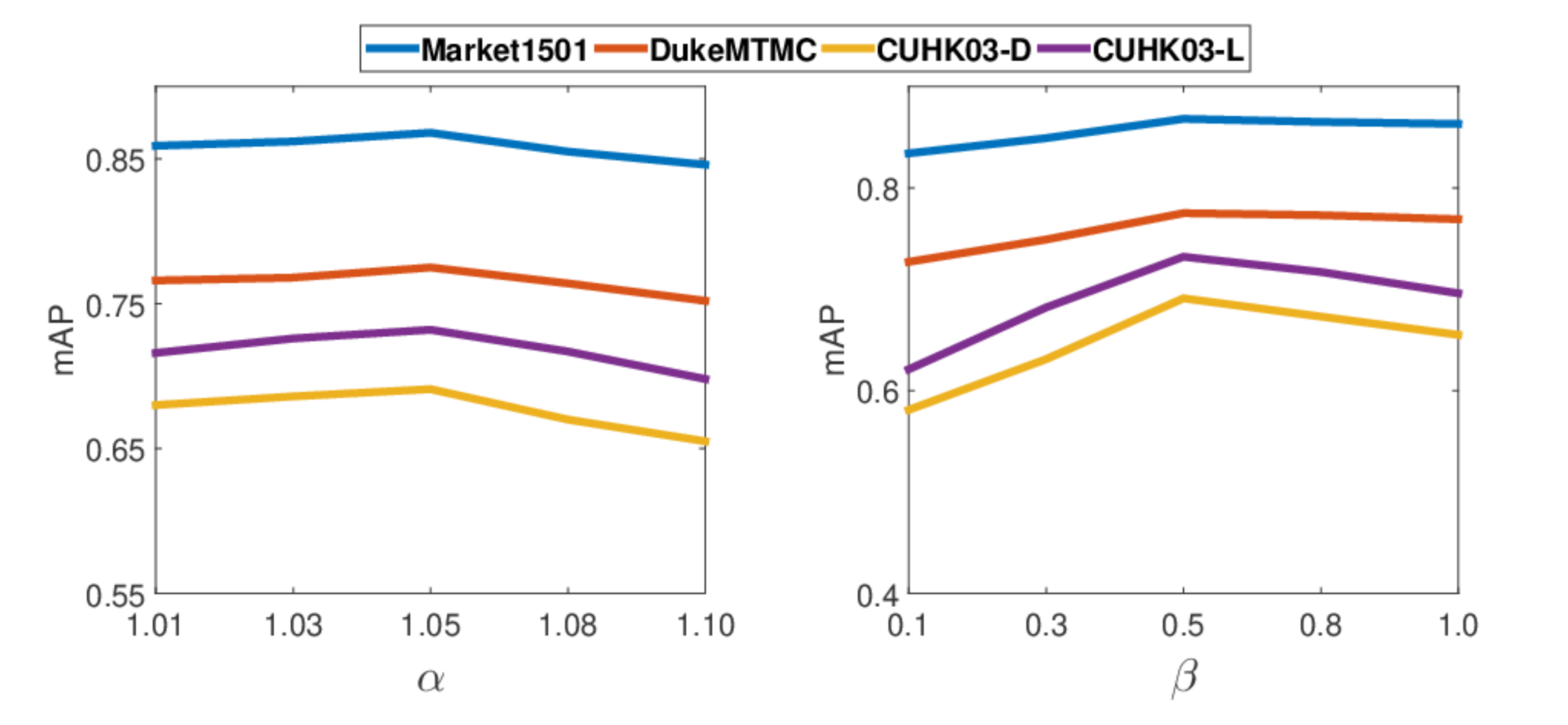}}
\end{minipage}

\caption{mAP results w.r.t $\alpha$ and $\beta$ in the FIDI loss}

\label{fig:parameter}
\end{figure}

The two hyperparameters $\alpha$ and $\beta$, which respectively control the loss bound for easy samples and the sensitivity of the FIDI loss w.r.t. the pairwise distance, can be well tuned via cross validation. This section aims at providing some starting points of the parameter tuning based on our empirical results. Here $\alpha=1.05$ and $\beta=0.5$ are used by default and we vary one parameter with the other one fixed to examine its impact on the performance. Due to the page limit, we only present the mAP results of Baseline2 (FIDI) in Figure~\ref{fig:parameter}.

In general, FIDI is not sensitive to $\alpha$ unless  it is too large. When the $\alpha$ is set to a small value, the punishment on images of small differences reduces, which decreases the final performance. On the right panel, we can see that it is beneficial to set a large $\beta$ for challenging datasets CUHK03-D and CUHK03-L since in such cases our loss becomes more sensitive to the pairwise distance. Our loss imposes exponentially larger penalization on images of small differences that results in performance improvement. On the other hand, a relatively small $\beta$ is more plausible for handling easier datasets like Market1501 and DukeMTM.

\section{Conclusions}
\label{sec:6}
This paper introduces a novel loss function called fine-grained difference-aware (FIDI) pairwise loss for the person ReID task. The FIDI loss not only ensures a similar inter-class margin as the triplet loss, but more importantly, also effectively penalizes images of both fine-grained and large appearance differences, especially on images of fine-grained differences. This delivers a significant improvement of three types of recent state-of-the-art ReID models in terms of both effectiveness and data efficiency. The improvement is particularly remarkable on complex datasets on which most current methods fail to work effectively. Also, our FIDI loss is simple and can replace the triplet loss as a plugin. All these characteristics make the FIDI loss a substantially more effective alternative to the widely-used triplet loss. We are performing large-scale studies to examine the applicability of replacing the triplet loss with the FIDI loss in other critical computer vision tasks.

\section{Acknowledgements}
Cheng Yan and Xiao Bai were supported by the NSFC project no. 61772057 and BNSF project no. 4202039.

\bibliographystyle{IEEEtran}
\bibliography{egbib}

\end{document}